\documentclass{article}
\PassOptionsToPackage{numbers, compress}{natbib}
\usepackage[final]{nips_2018}
\usepackage[utf8]{inputenc}
\usepackage[T1]{fontenc}   
\usepackage{hyperref}      
\usepackage{url}   
\usepackage{booktabs} 
\usepackage{amsfonts}  
\usepackage{nicefrac}  
\usepackage{microtype} 
\usepackage{enumitem}
\usepackage{graphicx}
\usepackage{amsmath}
\usepackage{amssymb}
\usepackage{floatrow}
\newfloatcommand{capbtabbox}{table}[][\FBwidth]

\title{Measuring Depression Symptom Severity from Spoken Language and 3D Facial Expressions}

\author{
    Albert Haque$^1$ \qquad Michelle Guo$^1$ \qquad Adam S Miner$^{2,3}$ \qquad Li Fei-Fei$^1$\\
    $^1$Department of Computer Science, Stanford University \\
    $^2$Department of Psychiatry and Behavioral Sciences, Stanford University\\
    $^3$Department of Health Research and Policy, Stanford University
}

\begin{document}
    
\maketitle

\begin{abstract}
With more than 300 million people depressed worldwide, depression is a global problem.
Due to access barriers such as social stigma, cost, and treatment availability, 60\% of mentally-ill adults do not receive any mental health services. Effective and efficient diagnosis relies on detecting clinical symptoms of depression. Automatic detection of depressive symptoms would potentially improve diagnostic accuracy and availability, leading to faster intervention.
In this work, we present a machine learning method for measuring the severity of depressive symptoms.
Our multi-modal method uses 3D facial expressions and spoken language, commonly available from modern cell phones.
It demonstrates an average error of 3.67 points (15.3\% relative) on the clinically-validated Patient Health Questionnaire (PHQ) scale.
For detecting major depressive disorder, our model demonstrates 83.3\% sensitivity and 82.6\% specificity.
Overall, this paper shows how speech recognition, computer vision, and natural language processing can be combined to assist mental health patients and practitioners.
This technology could be deployed to cell phones worldwide and facilitate low-cost universal access to mental health care.
\end{abstract}

\section{Introduction}
Worldwide, more than 300 million people are depressed \cite{who2018depression}.
In the worst case, depression can lead to suicide, with close to 800,000 people committing suicide every year.
In general, patients with mental disorders are seen by a wide spectrum of health care providers, including primary care physicians \cite{kroenke2001phq}.
However, compared to physical illnesses, mental disorders are more difficult to detect.
The burden of mental health is exacerbated by barriers to care such as social stigma, financial cost, and a lack of accessible treatment options.
To address entrenched barriers to care, scalable approaches for detecting mental health symptoms have been called for \cite{kazdin2011rebooting}.
If successful, early detection may impact access for the 60\% of mentally-ill adults who do not receive treatment \cite{nami}.

In practice, clinicians identify depression in patients by first measuring the severity of \textit{depressive symptoms}\footnote{Depressive symptoms include feelings of worthlessness, loss of interest in hobbies, or thoughts of suicide.} during in-person clinical interviews.
During these interviews, clinicians assess both verbal and non-verbal indicators of depressive symptoms including monotone pitch, reduced articulation rate, lower speaking volumes \cite{hall1995nonverbal, sobin1997psychomotor}, fewer gestures, and more downward gazes \cite{waxer1974nonverbal, schelde1998major, perez2003nonverbal}.
If such symptoms persist for two weeks \cite{american2013diagnostic}, the patient is considered to have a \textit{major depressive episode}. Structured questionnaires have been developed and validated in clinical populations to assess the severity of depressive symptoms. One of the most common questionnaires is the Patient Health Questionnaire (PHQ) \cite{kroenke2009phq}.
This clinically-validated tool measures depression symptom severity across several personal dimensions \cite{kroenke2002phq}. 
Assessing symptom severity is time-intensive, and critical for both initial diagnosis and improvement across time.
Thus, AI-based solutions to assessing symptom severity may address entrenched barriers to access and treatment.

We envision an AI-based solution where depressed individuals can receive evidence-based mental health services while avoiding existing barriers to access.
Such a solution could leverage multi-modal sensors or text messages, as is common on modern smartphones, to increase timely and cost-effective symptom screening \cite{althoff2016large}.
Conversational AIs are another potential solution \cite{miner2017talking, miner2016smartphone}.
Our hope is that automated feedback will (i) provide actionable feedback to individuals who may be depressed, and (ii) improve automated depression screening tools for clinicians, by including visual, audio, and linguistic signals.

\textbf{Contributions.}
We propose a machine learning method for measuring depressive symptom severity from de-identified multi-modal data.
The input to our model is audio, 3D video of facial keypoints, and a text transcription of a patient speaking during a clinical interview.
The output of our model is either a PHQ score or classification label indicating major depressive disorder.
Our method leverages a causal convolutional network (C-CNN) to ``summarize" sentences into a single embedding.
This embedding is then used to predict depressive symptom severity.
In our experiments, we show how our sentence-based model performs in relation to word-level embeddings and prior work.

\begin{figure*}[t]
\includegraphics[width=1.0\textwidth]{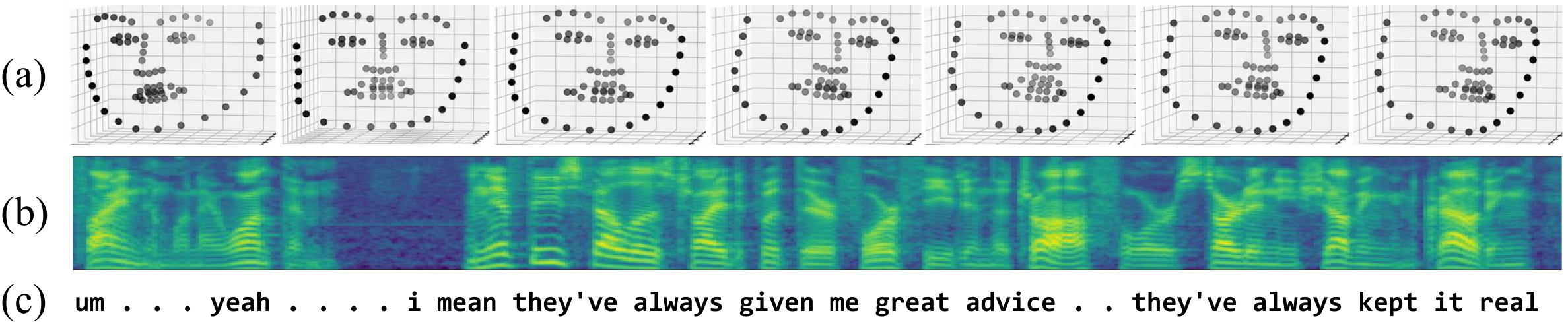}
\caption{\textbf{Multi-modal data.} For each clinical interview, we use: (a) video of 3D facial scans, (b) audio recording, visualized as a log-mel spectrogram, and (c) text transcription of the patient's speech. Our model predicts the severity of depressive symptoms using all three modalities.}
\label{fig:data}
\end{figure*}

\section{Dataset}
We use the DAIC-WOZ dataset \cite{gratch2014distress} containing audio and 3D facial scans of depressed and non-depressed patients.
For each patient, we are provided with the PHQ-8 score.
This corpus is created from semi-structured clinical interviews where a patient speaks to a remote-controlled digital avatar.
The clinician, through the digital avatar, asks a series of questions specifically aimed at identifying depressive symptoms.
The agent prompts each patient with queries that included questions (e.g. ``How often do you visit your hometown?") and conversational feedback (e.g. ``Cool.").
A total of 50 hours of data was collected from 189 clinical interviews from a total of 142 patients.
Following prior work \cite{al2018detecting}, results in our paper are from the validation set.
More details can be found in Appendix \ref{sec:appendix}.

\textbf{Privacy.}
Data used in this work does not contain protected health information (PHI).
Mentions of personal names, specific dates, and locations were removed from the audio recording and transcription by the dataset curators \cite{gratch2014distress}.
The 3D facial scans are low-resolution (68 pixels) and do not contain enough information to identify the individual but contain just enough to measure facial motions such as eye, lip, and head movements.
While the dataset is publicly available, future researchers who apply our method to other datasets may encounter PHI and should design their experiments appropriately.

\section{Model}

\begin{figure*}[t]
\includegraphics[width=1.0\textwidth]{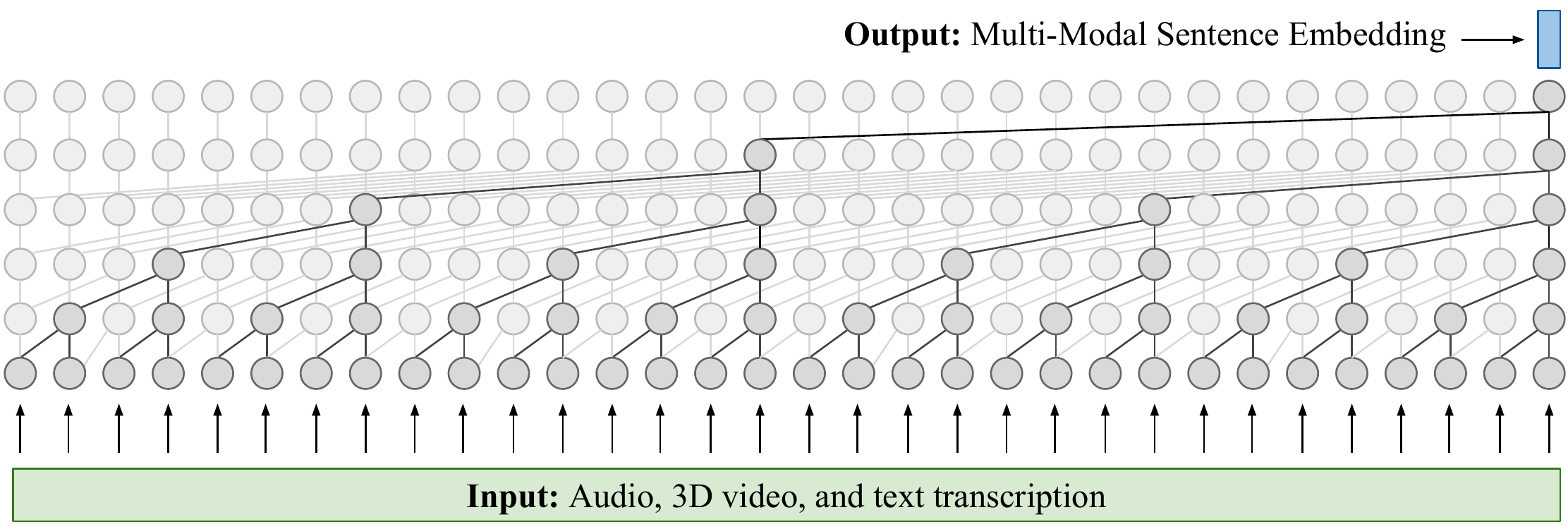}
\caption{\textbf{Our method: Learning a multi-modal sentence embedding.} Overall, our model is a causal CNN \cite{bai2018empirical}. The input to our model is: audio, 3D facial scans, and text. The multi-modal sentence embedding is fed to a depression classifier and PHQ regression model (not shown above).}
\label{fig:method}
\end{figure*}

Our model consists of two technical pieces: (i) a sentence-level ``summary" embedding and (ii) a causal convolutional network (C-CNN).
An overview is shown in Figure \ref{fig:method}.

\textbf{Sentence-Level Embeddings.}
For decades, word and phoneme-level embeddings\footnote{The goal of an \textit{embedding} is ``summarize" a variable-length sequence as a fixed-size vector of numbers.} have been the go-to feature for encoding text and speech \cite{bourlard1990continuous, bengio2014word, pennington2014glove, turian2010word}.
While these embeddings work well for some tasks \cite{sotelo2017char2wav, kim2016character, peters2018deep}, they are limited in their sentence-level modeling ability.
This is because word and phoneme embeddings capture a narrow temporal context, often a few hundred milliseconds at most \cite{waibel1990phoneme, labov200650}.
In this work, we propose a novel multi-modal sentence-level embedding.
This allows us to capture long-term acoustic, visual, and linguistic elements.

\textbf{Causal Convolutional Networks.}
During clinical interviews, patients may stutter and frequently pause between words.
This causes audio-video recordings to be longer than non-depressed patients.
Recently, causal convolutional networks (C-CNNs) have been shown to outperform recurrent neural networks (RNNs) on long sequences \cite{bai2018empirical}.
In \cite{miller2018recurrent}, the authors even show that RNNs can be approximated by fully feed-forward networks (i.e., CNNs).
Combined with dilated convolutions \cite{oord2016wavenet}, C-CNNs are well-poised to model the long sequences for depression screening interviews.
For a more thorough comparison of C-CNNs vs RNNs, we refer the reader to Bai et al. \cite{bai2018empirical}.

\section{Experiments}

Our experiments consist of two parts.
First, we compare our method to existing works for measuring the severity of depressive symptoms (Table \ref{tab:part1}).
We predict both the PHQ score and output a binary classification as to whether the patient has \textit{major depressive disorder}, typically with a PHQ score greater than or equal to ten \cite{manea2012optimal}.
Second, we perform ablation studies on our model to better understand the effect of multiple modalities and sentence-level embeddings (Table \ref{tab:part2}).
Data formats, neural network architectures, and key hyperparameters can be found in Appendix \ref{sec:appendix}.

\textbf{4.1 \quad Automatically Measuring the Severity of Depressive Symptoms}

\begin{table}[t]
\footnotesize
\begin{tabular}{@{}cl|c|ccc|c@{}}
\toprule
     & \multicolumn{1}{c}{} &  \multicolumn{4}{|c|}{Classification: Major Depressive Disorder} & \multicolumn{1}{c}{Regression: PHQ Score} \\ \midrule
     \# & Method & Modalities & F1 Score & Precision & Recall & Average Error \\ \midrule
     1 & SVM \cite{valstar2016avec} & A & 46.2 & 31.6 & 85.7 & 6.74 \\ 
     2 & CNN+LSTM \cite{ma2016depaudionet} & A & 52.0 & 35.0 & 100.0 & --- \\
     3 & SVM \cite{valstar2016avec} & V & 50.0 & 60.0 & 42.8 & 7.13 \\ 
     4 & Williamson et al. \cite{williamson2016detecting} & V & 53.0 & --- & --- & 5.33 \\
     5 & Williamson et al. \cite{williamson2016detecting} & L & 84.0 & --- & --- & 3.34 \\
     6 & Alhanai et al. \cite{al2018detecting} & AL & 77.0 & 71.0 & 83.0 & 5.10 \\ 
     7 & SVM \cite{valstar2016avec} & AV & 50.0 & 60.0 & 42.8 & 6.62 \\ 
     8 & Gong et al. \cite{gong2017topic} & AVL & 70.0 & --- & --- & 2.77 \\
     9 & Williamson et al. \cite{williamson2016detecting} & AVL & 81.0 & --- & --- & 4.18 \\
\midrule
     10 & C-CNN \cite{bai2018empirical} & AVL & 76.9 & 71.4 & 83.3 & 3.67 \\
     \bottomrule 
\end{tabular}
\caption{\textbf{Comparison of machine learning methods for detecting depression.} Two tasks were evaluated: (i) binary classification of major depressive disorder and (ii) PHQ score regression. Modalities: A: audio, V: visual, L: linguistic (text), AVL: combination. For prior work, numbers are reported from the original publications. Dashes indicate the metric was not reported. }
\label{tab:part1}
\end{table}

In Table \ref{tab:part1}, we compare our method to prior work on measuring depressive symptom severity.
One difference between our method and prior work is that our method does not rely on interview context.
Prior work depends heavily on interview context such as the type of question asked \cite{al2018detecting, gong2017topic}, whereas our method accepts a sentence without such metadata.
While additional context typically helps the model, it can introduce technical challenges such as having too few training examples per contextual class.
Another difference is that our method uses \textit{raw} input modalities: audio, visual, and text.
Prior work uses engineered features such as min/max vocal pitch and word frequencies.

\textbf{4.2 \quad Ablation Study}

In Table \ref{tab:part2}, rows 1-6 denote hand-crafted or pre-trained sentence-level embeddings.
That is, the entire input sentence (audio, 3D facial scans, and transcript) is summarized into a single vector \cite{mikolov2013distributed, le2014distributed, cer2018universal}.
However, we propose to learn a sentence-level embedding from the input.
These are shown in rows 7 and 8.
It is important to note that our method \textit{does} use hand-crafted and pre-trained \textit{word}-level embeddings as input.
However, internally, oure model learns a \textit{sentence}-level embedding.
Following prior work on sentence-level embeddings, rows 1-6 were computed via simple average \cite{cer2018universal}.
To learn sentence-level embeddings, we evaluate: (i) long short-term memory \cite{hochreiter1997long} and (ii) causal convolutional networks \cite{oord2016wavenet, bai2018empirical}.

\begin{table}[t]
\footnotesize
\begin{tabular}{@{}rl|cccc|cccc|cccc@{}}
\toprule
 &  & \multicolumn{8}{c|}{Classification: Major Depressive Disorder} & \multicolumn{4}{c}{Regression: PHQ Score} \\ \midrule
 &  & \multicolumn{4}{|c|}{Sensitivity (TPR)} & \multicolumn{4}{c|}{Specificity (TNR)} & \multicolumn{4}{c}{Average Error} \\ \midrule
\# & Method & A & V & L & AVL & A & V & L & AVL & A & V & L & AVL \\ \midrule
1 & Log-Mel & 70.5 & --- & --- & --- & 64.5 & --- & --- & --- & 6.40 & --- & --- & --- \\
2 & MFCC & 74.3 & --- & --- & --- & 67.7 & --- & --- & --- & 5.96 & --- & --- & --- \\
3 & 3D Face \cite{openface} & --- & 71.4 & --- & --- & --- & 69.4 & --- & --- & --- & 5.82 & --- & --- \\
4 & W2V \cite{mikolov2013distributed} & --- & --- & 65.3 & --- & --- & --- & 64.6 & --- & --- & --- & 6.16 & --- \\
5 & D2V \cite{le2014distributed} & --- & --- & 67.7 & --- & --- & --- & 71.4 & --- & --- & --- & 5.81 & --- \\
6 & USE \cite{cer2018universal} & --- & --- & 63.4 & --- & --- & --- & 62.5 & --- & --- & --- & 6.27 & --- \\ \midrule
7 & LSTM \cite{hochreiter1997long} & 53.9 &  61.0 & 57.5 & 74.2 & 54.6 & 59.7 & 60.3 & 76.1 & 7.15 & 6.12 & 6.57 & 5.18 \\
8 & C-CNN \cite{bai2018empirical} & 71.1 & 73.7 & 67.7 & 83.3 & 66.7 & 71.2 & 65.5 & 82.6 & 5.78 & 5.01 & 6.14 & 3.67 \\ \bottomrule
\end{tabular}
\caption{\textbf{Ablation study.}  Rows 1-2 are hand-crafted embeddings. Rows 3-6 are pre-trained embeddings. Rows 7-8 denote our learned sentence-level embeddings. Modalities: A: audio, V: visual, L: linguistic (text), AVL: combination. TPR and TNR denote true positive and negative rate, respectively. The input to 7-8 were sequences of log-mel spectrograms, 3D faces, and Word2Vecs.}
\label{tab:part2}
\end{table}

\section{Discussion}

Before adapting our work to future research, there are some points to consider.
First, although a human was controlling the digital avatar, the data was collected from human-to-\textit{computer} interviews and not human-to-\textit{human}.
Compared to a human interviewer, research has shown that patients report lower fear of disclosure and display more emotional intensity when conversing with an avatar \cite{lucas2014s}.
Additionally, people experience psychological benefits from disclosing emotional experiences to chatbots \cite{ho2018psychological}.
Second, although it is commonly used in treatment settings and clinical trials, the symptom severity score (PHQ) is not the same as a formal diagnosis of depression.
Our work is meant to augment existing clinical methods and not issue a formal diagnosis.
Finally, while pre-existing embeddings are easy to use, recent research suggests these vectors may contain bias due to the underlying training data \cite{caliskan2017semantics, bolukbasi2016man, garg2018word}.
Mitigating bias is outside the scope of our work, but is crucial to providing culturally sensitive diagnosis and treatment.

Future work could better utilize longitudinal and temporal information such as depression scores across interview sessions that are weeks or months apart.
Understanding \textit{why} the model made certain predictions could also be valuable.
Visualizations such as confidence maps over the 3D face and ``usefulness" scores for audio segments could shed new insights.

In conclusion, we presented a multi-modal machine learning method which combines techniques from speech recognition, computer vision, and natural language processing.
We hope this work will inspire others to build AI-based tools for understanding mental health disorders beyond depression.

\textbf{Acknowledgements.} This work was supported by a National Institutes of Health, National Center for Advancing Translational Science, Clinical and Translational Science Award (KL2TR001083 and UL1TR001085).
The content is solely the responsibility of the authors and does not necessarily represent the official views of the NIH.

\clearpage
\newpage
{
\setlength{\bibsep}{4pt plus 0.3ex}
\bibliographystyle{abbrv}
\bibliography{references}
}

\clearpage
\newpage
\appendix
\section{Appendix}\label{sec:appendix}

\subsection{Data Format}\label{sec:data_format}
Full data details can be found on the original dataset website \cite{baltruvsaitis2016openface}.
Audio was recorded with a head-mounted microphone at 16 kHz.
Video was recorded at 30 frames per second with a Microsoft Kinect.
A total of 68 three-dimensional facial keypoints were extracted using OpenFace \cite{baltruvsaitis2016openface}.
Audio was transcribed by the dataset curators and segmented into sentences and phrases with millisecond-level timestamps \cite{gratch2014distress}.
We use the dataset's train-val split: train (107 patients), validation (35 patients).
Note that while a test set exists, the labels are not public.

We canonicalized slang words present in the transcription. For example, \textit{bout} was translated to \textit{about}, \textit{till} was translated to \textit{until}, and \textit{lookin} was translated to \textit{looking}.
All text was forced to lower case.
Numbers were canonicalized as well (e.g., 24 was represented as \textit{twenty four}).

\subsection{Implementation Details}

\subsubsection{Experiment 1: Automatically Measuring the Severity of Depressive Symptoms}\label{sec:appendix_exp1}

Input to ``our method", i.e. Causal CNN are as follows:
\begin{itemize}[itemsep=0pt,topsep=0pt,leftmargin=20pt]
    \item Audio: Log-mel spectrograms with 80 mel filters.
    \item Visual: 68 3D facial keypoints.
    \item Linguistic: Word2Vec embeddings \cite{mikolov2013distributed}.
\end{itemize}

The network architecture is a 10-layer causal convolutional network \cite{bai2018empirical} with kernel size of 5 with 128 hidden nodes per layer.
Dropout was applied to all non-linear layers with a 0.5 probability of being zeroed.
The loss objectives were binary cross entropy for classification and mean squared error for regression.
The model was optimized with the Adam optimizer with $\beta_1 = 0.9$ and $\beta_2 = 0.999$ with L2 weight decay of 1e-4.
The initial learning rate was 1e-3 for classification and 1e-5 for regression.
A batch size of 16 was used.
The model was trained on a single Nvidia V100 GPU for 100 epochs.
Our model was implemented with Pytorch \cite{paszke2017automatic}.

\subsubsection{Experiment 2: Ablation Studies}

For Table \ref{tab:part2}, the details for each row are as follows:
\begin{enumerate}[itemsep=0pt,topsep=0pt,leftmargin=20pt]
    \item Log-mel spectrograms were computed with 80 mel filters.
    \item Mel-frequency cepstral coefficients were computed with 13 resulting values.
    \item A total of 68 three-dimensional facial keypoints were provided by the dataset \cite{gratch2014distress}. They were extracted using OpenFace \cite{openface}.
    \item Word2Vec vectors were computed using the publicly available Word2Vec model from Google and the Gensim python library \cite{rehurek_lrec}. Each vector is of length 300.
    \item Doc2Vec vectors were also computed using Gensim \cite{rehurek_lrec}. Each vector is of length 300.
    \item Universal sentence embeddings were computed using Tensorflow \cite{abadi2016tensorflow} from the publicly available release. Each vector is of length 512.
    \item The LSTM consists of 10 layers with 128 hidden units and is also optimized with the same batch size, optimizer, etc. as stated in Appendix \ref{sec:appendix_exp1}.
    \item Our causal CNN model is the same as the one outlined in Appendix \ref{sec:appendix_exp1}.
\end{enumerate}
Public code implementations\footnote{\url{https://github.com/locuslab/TCN}} were used for the core network architecture components for both the LSTM and causal CNN.
\end{document}